# Hybrid-DNNs: Hybrid Deep Neural Networks for Mixed Inputs

Zhenyu Yuan[1*], Yuxin Jiang[2], Jingjing Li[3], Handong Huang[1]

[1] *China University of Petroleum-Beijing, Beijing, China.*

[2] *PST Service Corporation, Beijing, China.*

[3] *Beijing Power Concord Technology Co. Ltd., Beijing, China.*

**Abstract**

Rapid development of big data and high-performance computing have encouraged explosive studies of deep learning in geoscience. However, most studies only take single-type data as input, frittering away invaluable multi-source, multi-scale information. We develop a general architecture of hybrid deep neural networks (HDNNs) to support mixed inputs. Regarding as a combination of feature learning and target learning, the new proposed networks provide great capacity in high-hierarchy feature extraction and in-depth data mining. Furthermore, the hybrid architecture is an aggregation of multiple networks, demonstrating good flexibility and wide applicability. The configuration of multiple networks depends on application tasks and varies with inputs and targets. For reservoir production prediction, a specific HDNN model is configured. Considering their contributions to hydrocarbon production, core photos, logging images and curves, geologic and engineering parameters can all be taken as inputs. After preprocessing, the mixed inputs are prepared as regular-sampled structural and numerical data. For feature learning, convolutional neural networks (CNN) and multilayer perceptron (MLP) network are configured to separately process structural and numerical inputs. Learned features are then concatenated and fed to subsequent networks for target learning. Comparison with typical MLP model and CNN model highlights the superiority of proposed HDNN model with high accuracy and good generalization.

**Keyworks**

Deep Learning; Hybrid Networks; Deep Neural Networks; CNNs; Mixed Inputs; Geoscience

**Highlights**

- We develop a novel architecture for hybrid deep neural networks (HDNNs) to support mixed inputs.
- Mixed inputs indicate data of various types or formats, provide more aspects of features.
- The combination of feature learning and target learning provides great capacity in high-hierarchy feature extraction and in-depth data mining.
- The aggregation of multiple networks demonstrates good flexibility and wide applicability.
- The application to reservoir production prediction highlights the accuracy and generalization of proposed HDNNs.

\* Corresponding author. zhenyuyuan@outlook.com.

## 1. Introduction

With the rapid development of big data and high-performance computers, it is achievable to extract more information and gain new insights from extensive datasets. Techniques from the rapidly evolving field of machine learning play a key role in this effort. Machine learning provides scientists with a set of tools for discovering new patterns, structures, and relationships in scientific datasets that are not easily revealed through conventional techniques (Bergen et al., 2019). Nowadays, machine learning is widely applied in various industries to facilitate tasks such as data analysis, pattern recognition, target prediction, and so on. For oil and gas industry, machine learning techniques were introduced to help geoscientists and engineers answering persistent questions about how to locate and develop economic hydrocarbon resources. Bergen et al. (2019) have reviewed applications of machine learning for data-driven discovery in solid earth geoscience. Focus on petrophysics, Xu et al. (2019) investigated the capacity and performance of machine learning dealing with big data.

Efforts to understand the solid earth are challenged by the fact that nearly all of earth's interior is, and remains, inaccessible to direct observation (Bergen et al., 2019). Instead, knowledge of interior properties and processes are based on measurements taken at or near the surface and discovered by solving inverse problems connecting measurements and targets. Due to the heterogeneity and complexness of sedimentary deposits and limitation of measurements, the solutions of these inverse problems are often indeterminate (Koltermann and Gorelick, 1996). However, the largest obstruction is not from our inability to solve the equations, but from knowing what the interior structure of the earth is really like and the parameters that should go into those equations (Bergen et al., 2019). Theoretical knowledge is still incomplete, lots of tasks are difficult for humans to perform or explain. Commonly used physics-driven methods are often assumptions-based and data-restricted, restraining their applicability and generalization. In comparison, machine learning takes advantage of big data and can excavate complex relationships between measurements and observations. Therefore, it is well suited to address those problems.

Over the past decade, the amount of data available to geoscientists has grown enormously, through larger deployments of traditional sensors and through new data sources and sensing modes (Bergen et al., 2019). Xu et al. (2019) have summarized typically acquired data from various sources in petroleum industry, including core measurements, wellbore measurements, remotely sourced measurements and reservoir performances. These data can be categorized into different categories such as geological, geophysical, petrophysical, or reservoir engineering. Furthermore, the types of above data include numerical value, category, image, text and so on. In this "big data" world, we're often presented with an abundance of features that could be used to do machine learning. Some will be more useful than others, and some will be basically useless noise. Even if the data consists of only a few features, we may have found that two or more are highly correlated. In a situation like this, it's a common practice to skip other correlative features and use only one for modeling purposes. In problems where dozens, hundreds, or even thousands of possible features exist, statistical techniques are usually utilized to decide which features are the most important. Taking a machine learning regression application as example, Yuan et al. (2018b) performed feature representing under principles of high contribution, good consistency and strong orthogonality, through single-attribute analysis and multi-attribute analysis. Instead of explicit statistical techniques, appropriate architectures involved in deep neural networks (DNNs) can effectively extract features and their corresponding weights through representation learning (Bengio et al., 2013). As a subfield of machine learning, deep learning, usually by DNNs, uses multiple layers to progressively extract higher-level features from raw inputs (Deng and Yu, 2014).

During the past few years, deep learning has been introduced in various aspects of geoscience applications. Here we highlight some of them in exploration geophysics and petrophysics. For exploration geophysics, deep learning techniques have been introduced into seismic processing for first-break picking (Duan et al., 2018; Hu et al., 2019; Yuan et al., 2018a), data regularization (Lu et al., 2019a; Wang et al., 2019b), denoising and image enhancement (Dong et al., 2019; Dutta et al., 2019; Halpert, 2018; Siahkoohi et al., 2019; Sun and Demanet, 2018; Sun et al., 2019; Zhang et al., 2019a; Zhang et al., 2019b), velocity modeling (Li et al., 2018; Park and Sacchi, 2019; Wang and Ma, 2019; Wu and Lin, 2019) and imaging (Herrmann et al., 2019). In addition, there were some seismic interpretation studies, including fault detection (Huang et al., 2017; Wu et al., 2019a; Wu et al., 2019b; Wu et al., 2019c; Xiong et al., 2018; Yuan et al., 2019), seismic facies segmentation (Duan et al., 2019; Krasnov et al., 2018; Mukhopadhyay and Mallick, 2019; Pham et al., 2019; Titos et al., 2019; Waldeland et al., 2018; Zhao, 2018; Zhao et al., 2016), automatic horizon picking (Di et al., 2019; Yang and Sun, 2019) as well as seismic inversion (Biswas et al., 2019; She et al., 2019; Wang et al., 2019a). In the petrophysics discipline, there were deep learning applications for permeability prediction (Zhong et al., 2019), reservoir thickness estimation (Lu et al., 2019b) and lithology facies recognition (Jaikla et al., 2019; Zhang et al., 2018). Above studies have made considerable progresses in certain tasks, but they commonly consider only one single type of data as input. In practice, multiple types of data may contribute to the performance of target. Therefore, it is preferable to take more types of measurements into consideration for deep learning. Accordingly, some advanced network architectures are required.

It is well accepted that CNNs play an important role in learning excellent features for image processing tasks. However, in tradition they only allow adjacent layers connected, limiting integration of multi-scale or mixed-type information. Li et al. (2017) presented a framework concatenating multi-scale features by shortcut connections to the latter fully-connected layer and achieved better results than traditional convolutional neural networks (CNNs) on image classification and recognition. Considering the target features submerging in complex background loads, Wu and Wang (2019) developed a concatenate convolutional neural network to separate the feature of the target load from the load mixed with the background. In this paper, we develop a type of hybrid architecture to construct DNNs to handle mixed inputs. In the following sections, we will firstly introduce the general architecture of proposed hybrid deep neural networks (HDNNs) and some relevant theories. Then a specific network for reservoir production prediction is presented and its application to a practical survey is demonstrated. Finally, we draw some conclusions and give some suggestions.

## 2. Architecture of Hybrid Deep Neural Networks

### 2.1 General architecture

In common, things are interconnected with each other and the output performance is determined by more than one single factor. Taking multiple factors into consideration, a general architecture of HDNNs is proposed to evaluate the influence of each factor to the target. The general architecture is shown in Figure 1, which can be treated as a combination of two parts, namely feature learning and target learning.

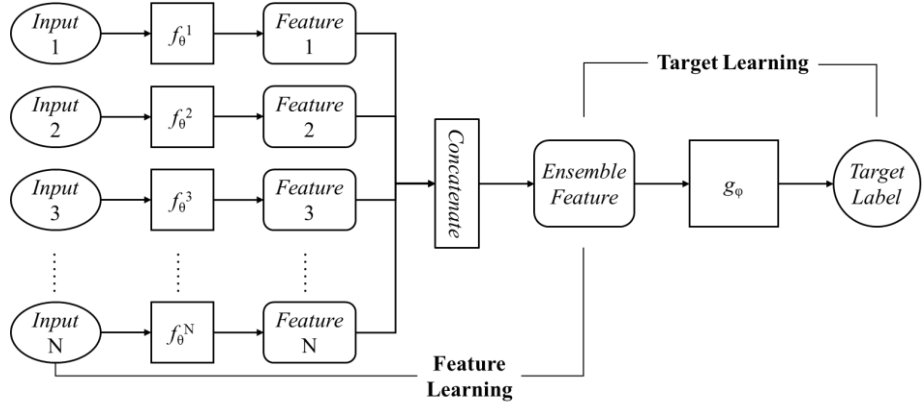

**Figure 1** *General architecture of hybrid deep neural networks, where f and g denote nonlinear equations defined by certain neural networks, θ and φ indicate corresponding weights learned by deep learning, number varying from 1 to N indicates index of input data.*

Multiple inputs from different measurements are usually of different types or formats. It is not available or convenient to input all these data into a single network. The proposed HDNNs separately handle each input with an appropriate network to perform feature learning. Then all learned features are concatenated into an ensemble feature. This ensemble feature contains valid information from different inputs and then is fed to subsequent neural network to perform target learning.

As a general framework to handle mixed-input tasks, the inputs and neural networks are not restricted. For different applications, the inputs can be video, image, audio, text, numerical value or categorical tag. The dimension of input data varies from point, 1D, 2D, 3D to even much higher. What's more, the sampling index of each input could be either continuous or discrete. To handle various inputs, corresponding neural networks could be multiple-layer perceptron (MLP), CNNs, recurrent neural networks or their multiple variations. Furthermore, the so-called neural networks could be traditional machine learning algorithms as well, such as support vector machine, random forests or others.

### 2.2 Network representation

Considering N types of data as inputs, the mapping equation from inputs to learned features is expressed as

$$\mathbf{T}^i = f_\theta{}^i(\mathbf{X}^i),\tag{1}$$

where $\mathbf{X}^i$ and $\mathbf{T}^i$ indicate certain type of input and its learned feature, corresponding network algorithm is denoted as $f_\theta{}^i$, $\theta$ indicates model weights, $i$ varies from 1 to N. It should be noted that the algorithm $f_\theta{}^i$ varies according to the type or format of $\mathbf{X}^i$ and $\mathbf{T}^i$. For instance, if the input data is a 1D series, the corresponding algorithm could be 1D CNN or RNN. While if the input data is image or video, the corresponding algorithm should better be 2D or 3D CNN.

Concatenating all learned features, an ensemble feature $\mathbf{Z}$ is achieved.

$$\mathbf{Z} = \mathrm{concat}\,(\mathbf{T}^1, \mathbf{T}^2, \ldots, \mathbf{T}^N),\tag{2}$$

where "concat" indicates concatenate operation. The concatenate operation could be channel-wise or feature-wise,

depending on the formats of inputs. As an illustration, if there were inputs of point dimension, we should better concatenate the learned features in the feature-wise way.

After concatenate operation, the ensemble feature is taken as input to the target learning model $g_\varphi$. Since the target label can be either continuous value or categorical class, the target learning model applies to both regression and classification applications. Indicating target label as $Y$, it is derived from

$$Y = g_\varphi(\mathbf{Z}) \, . \tag{3}$$

Based on above derivation, the proposed HDNNs realize an end-to-end deep learning model, expressed as

$$Y = F(\mathbf{X}^1, \mathbf{X}^2, ..., \mathbf{X}^N) \, , \tag{4}$$

where $F$ is an integrated function representing the nonlinear mapping from multiple inputs to target label. It is a combination of $f_{\theta^i}$, concatenate operation and $g_\varphi$. Model weights $\theta$ and $\varphi$ are determined by solving optimization problem through deep learning.

*2.3 Optimization expression*

A loss function is requisite to perform optimization and varies according to the category of task. For regression application, mean squared error (MSE) loss function is commonly adopted, expressed as

$$MSE = \frac{1}{M} \sum_{j=1}^{M} \left\| F(\mathbf{X}_j^1, \mathbf{X}_j^2, ..., \mathbf{X}_j^N) - Y_j \right\|^2 \, , \tag{5}$$

where $j$ indicates the instance index, varying from 1 to $M$.

For classification application, the cross-entropy measure is taken as an example of loss function,

$$CE = -\sum_{j=1}^{M} \sum_{k=1}^{C} Y_{jk} \log(p_{jk}) \, , \tag{6}$$

where $k$ indicates class index, varying from 1 to C. $Y_{jk}$ indicates binary indicator of class $k$ for instance $j$, $p_{jk}$ indicates predicted probability of class $k$ for instance $j$. The calculation of probability is subject to the selection of $F$.

## 3. Application on Reservoir Production Prediction

For oil and gas industry, there are multiple sources of measurements, such as core measurements, wellbore measurements, remotely sourced measurements and reservoir performances, contributing to the discovery and evaluation of hydrocarbon resources. To evaluate the potential of underground reservoirs, hydrocarbon production is often regarded as one essential property. Only if the productivity is higher enough to cover the exploitation cost, a reservoir is economically exploitative.

*3.1 Related works on production prediction*

To predict hydrocarbon production, many researches have been done considering not only the storage and permeability of reservoir formations (Cheng et al., 1999; Hogg et al., 1996; Liu et al., 2000), but also the engineering factors such as hydraulic fracturing (Chen et al., 2019; Huang et al., 2015). However, physics-based methods often apply to certain type of reservoirs and require in-depth geological understanding to achieve better predictions. Taking advantage of machine learning, Pan et al. (2015) and Hu et al. (2018) have introduced neural networks into production prediction. It should be noted that production is a criterion of reservoir formation. However, the aforementioned methods took average values of log curves as inputs, discarding structural characteristics of reservoir formations. In contrast, our new developed HDNNs architecture is quite appropriate for production prediction.

*3.2 Specific HDNN architecture*

Measurements including cores, well logging, well test and engineering operations may have influences on reservoir productivity. The data types of above measurements include images, sequences, numerical values and categorical tags. To be specific, image data could be image logging (i.e. formation microimager (FMI)), core photos or scanning electron microscopy images. Well logging provides a variety of sequential curves. Some engineering parameters are numerical values, for example propping agent volume in hydraulic fracturing. Well test or geologic analysis provides categorical tags such as fluid type, lithology type or facies description. The sampling presentation of various data is also different. Numerical values and categorical tags are discrete, representing integral effect of reservoir formations. While images and sequences are presented as structural data, reflecting more details within the formations.

Taking all these mixed data as inputs, a specific HDNN architecture (shown as Figure 2) for production prediction is proposed. This architecture is represented as a combination of MLP and CNN. Specifically, MLP is adopted to deal with numerical and categorical inputs, CNN is applicable to extract high-hierarchy features from structural data. MLP consists of multiple fully connected (FC) layers. CNN is composed of several convolutional network units. After feature leaning for different formats of data seperately, outputs of MLP and CNN are concatenated feature-wise to achieve a final evaluation of production.

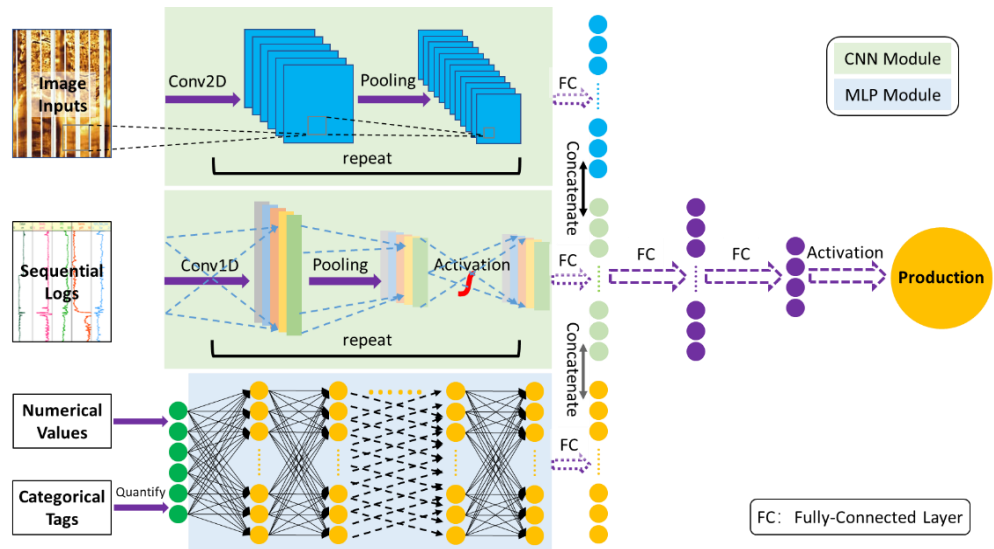

**Figure 2** *Architecture of HDNN for productivity prediction, referred to Yuan et al. (2020).*

Data preparation and preprocessing are essential for deep-learning production prediction. The inputs to HDNN should be firstly gathered from various wells. Then some preprocessing steps are taken to prepare well-sampled dataset. For categorical data, one-hot encoding is utilized to perform quantitative transformation. Log curves are extracted according to the depth range of each reservoir formation of each well, so do FMI images and core photos. Since the thicknesses of various formations are generally different, resampling is required to obtain uniform-sampled data. Furthermore, numerical and structural data are normalized to avoid the effect of inconsistent scales. Besides features, hydrocarbon production of each formation is set as target label. Finally, the prepared dataset is separated into training data and test data for model training and performance evaluation.

Through deep learning, the HDNN model for production prediction can be well trained, revealing the complex relationship between various measurements and target production. Then the well-trained model is applied to pre-prepared test data to evaluate its generalization performance. Finally, possible productions of some unknown wells can be predicted, and further guiding well location deployment and development engineering.

*3.3 Data preparation and analysis*

The proposed HDNN model is applied to an oil development block for production prediction. The block contains 180 development wells. Seven types of log curves, namely caliper, acoustic time, gamma ray, spontaneous potential, shale content, deep and shallow lateral resistivity, are available for all wells. Initial oil production is set as target label. Besides log curves, some formation and enigneering attributes related to initial production are also taken into consideration. These attributes include formation thickness, formation median depth, perforation thichness and perforation number. Table 1 shows available attributes and their corresponding data types. Representations of these attributes for a example formation are also demonstrated.

*Table 1 Features for deep learning productivity prediction.*

| Type | Numerical Data | | | | Sequential Log Curves | | | | | | |
|---|---|---|---|---|---|---|---|---|---|---|---|
| **Attribute Name** | Formation Thickness /m | Formation Depth /m | Perforation Thickness /m | Perforation Number | CAL | AC | GR | LLD | LLS | SP | VSH |
| **Example Attribute Presentation** | 96.1 | 2355.95 | 15.9 | 5 | 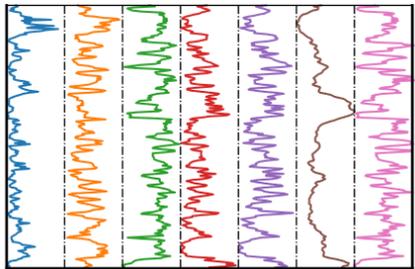 | | | | | | |

Figure 3 shows the relationship of formation and perforation attributes with oil production. For log curves, average values of curves within certain reservoir formation are computed to generally inspect their correlation with target production, shown in Figure 4. From both Figure 3 and Figure 4, we know that these discrete attributes have weak correlation to target oil production. That is to say, it is not practicable to predict produciton from discrete numerical attributes. In contrast, further incoporating structral log curves by HDNNs considers the spatial correlation and variation of reservoir formations, thus may enhance performance of production prediction.

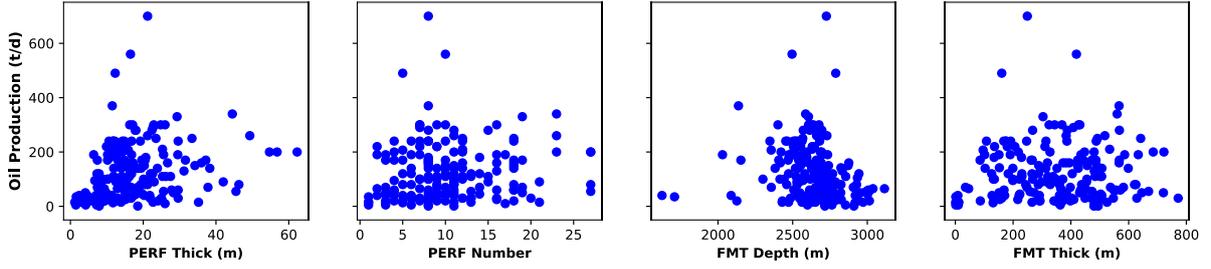

*Figure 3* Crossplots of oil production varying with perforation thickness, perforation number, formation depth and formation thickness.

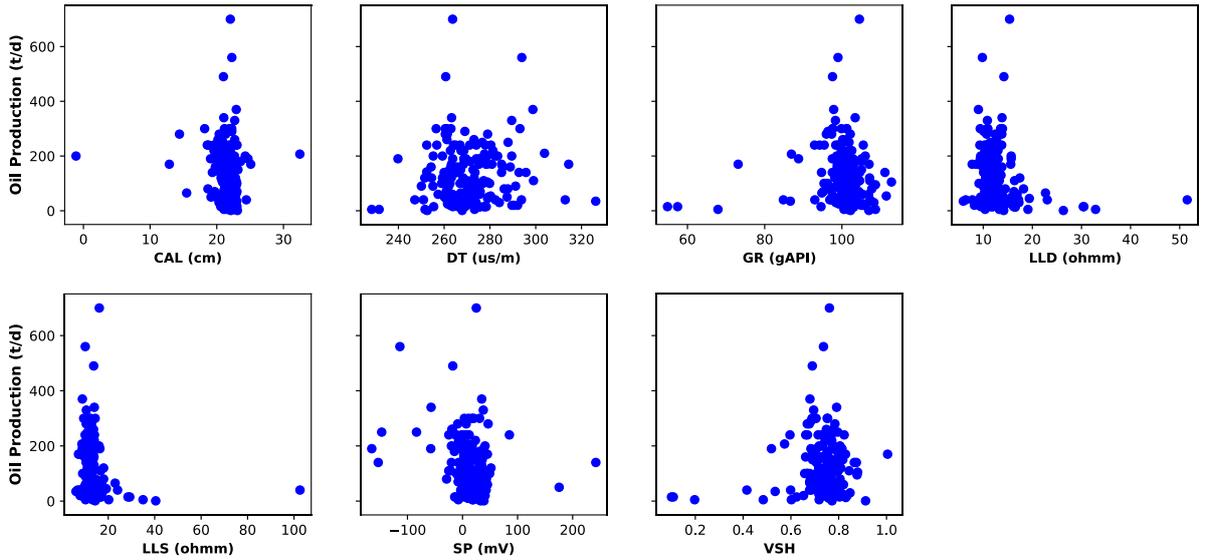

*Figure 4* Crossplots of oil production varying with average values of log curves within specific reservoir formation. Log curves are caliper, acoustic time, gamma ray, deep lateral resistivity, shallow lateral resistivity, spontaneous potential and shale content in sequence.

### 3.4 Model configuration and training

Due to the restriction of available measurements, only discrete numerical values and sequential log curves are provided here for initial oil production prediction. A customized model with only 1D CNN and MLP networks is adopted, taking advantage of both discrete numerical values and sequential log curves. Advanced regularization techniques such as batch normalization (Ioffe and Szegedy, 2015) and dropout (Srivastava et al., 2014) are adopted to facilitate deep learning. Batch normalization (BN) draws its strength from performing normalization for each training mini-batch. Dropout is efficient for reducing overfitting by randomly dropping units from the neural network during training. Both techniques are appropriate for improving the speed, performance, and stability of deep neural networks. Rectified linear units (ReLU), which enables better training of deeper networks, is used as an activation function.

In addition to the HDNN model, an MLP model and a CNN model are also adopted to make comparison. The MLP and CNN models share same architectures with the assembled MLP and CNN modules in HDNN. For MLP or CNN model training, single type of data is taken as input, and the corresponding network module is directly connected to the last FC layers to achieve output, without concatenate operation. For the MLP model, 11 numerical attributes are prepared as inputs, where sequential log curves are averaged as additional seven attributes. During model training, same optimizer (adaptive moment estimation (Kingma and Ba, 2014)) and loss function (MSE) are adopted to perform deep-learning optimization.

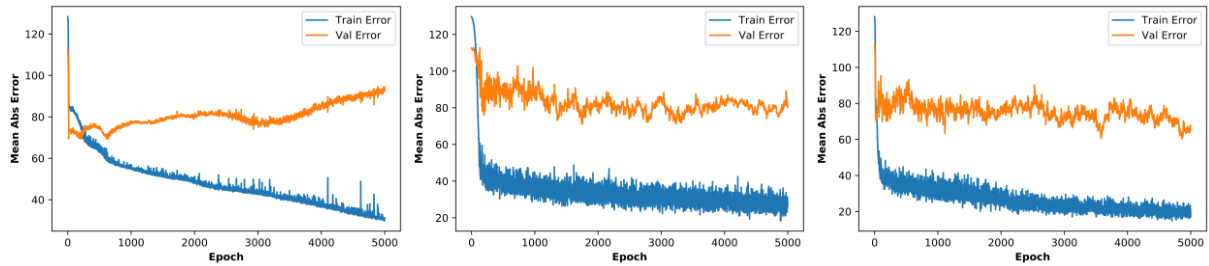

*Figure 5* Training performance of MLP model, CNN model and HDNN model in sequence.

Figure 5 shows the training performance of MLP, CNN and HDNN model for production prediction. The performance of MLP model indicates that it runs into overfitting, which may be caused by low correlation of features to target label and relatively deeper network layers. However, CNN model and HDNN model both show better convergence, with training error and validation error declining. Comparatively, HDNN model exhibits smoother fluctuation, better validation error descent and lower mean absolute error. That is to say, considering comprehensive mixed inputs, HDNN model performs better than typical CNN model.

Following model training, some test data are utilized to evaluate the model's performance. As showed in Figure 6, measured and predicted oil productions from three DNN models are demonstrated in crossplots. Meanwhile, squared correlation coefficient ($r^2$) is adopted as quantitative criterion and displayed. In accordance with the analyses from training performance, the HDNN model is superior to the typical CNN and MLP model, demonstrating best correlation with highest $r^2$. Rather than taking statistic averages as inputs, both CNN and HDNN model handle log curves with convolution operation and present good accuracy and generalization. It can be concluded that structural log curves contribute significantly to the hydrocarbon production.

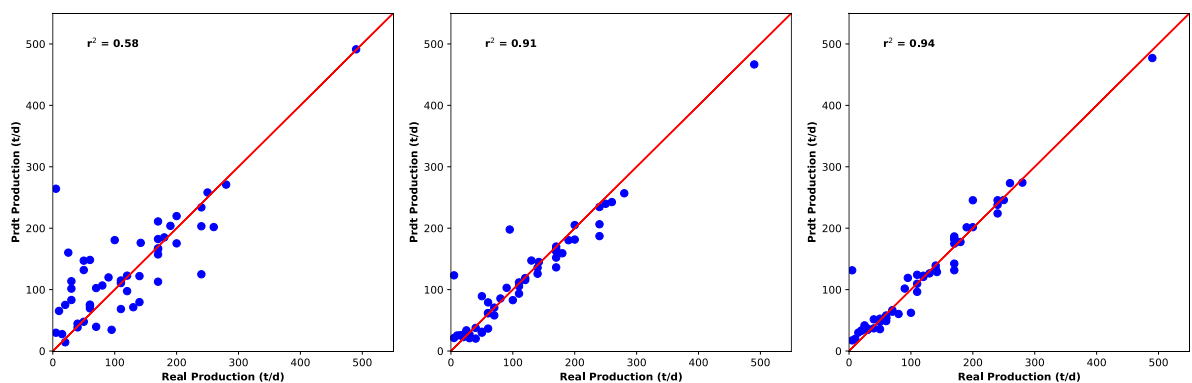

*Figure 6* Crossplots of predicted production versus measured production for MLP, CNN and HDNN model.

## 3.5 Production prediction

The well-trained HDNN model is finally used to three new wells to evaluate their production potentials. Three new wells are deployed and drilled based on available geologic evaluation, well logging is also implemented. From above analyses, we know structural log curves indicate possible geological capacity, while engineering parameters like perforation settings impact its presentation. Setting probable target formation and corresponding perforation parameters, the well-trained HDNN model is adopted to predict possible production. Meanwhile oil test engineering is performed. Information including predicted and measured initial oil productions are showed in Table 2. Though the predicted productions are a bit larger than the measured ones, the predictions basically indicate the potentials of each formation of various wells.

*Table 2* *Predicted and measured oil production of three new wells.*

| Well Name | Formation Range /m | Perforation Thickness /m | Perforation Number | Predicted Production t/d | Tested Production t/d |
|---|---|---|---|---|---|
| W1 | 2056.1-2058.8 | 2.7 | 2 | 33 | 20 |
| W2 | 1669.2-1676.5 | 4 | 4 | 120 | 105 |
| W3 | 1665.8-1997.5 | 3.7 | 3 | 20 | 5 |

## 4. Conclusions

Taking mixed data from various sources as inputs, we developed a general architecture of HDNNs. The consideration of mixed data takes full advantage of big data and is suitable for discovering more accurate relationship between measurements and target. In addition, the proposed HDNN model realizes an end-to-end learning, avoiding tedious works such as feature extraction and selection. The general HDNNs can be regarded as an aggregation of multiple network modules. The configuration of these modules depends on target application and varies with the mixed inputs and target labels. This innovative architecture provides HDNNs with good flexibility and wide applicability. The principle that multiple factors contribute to an outcome and the availability of diverse measurements support the versatility of proposed HDNNs for diverse deep learning applications. Concentrating on hydrocarbon production prediction, the HDNN model takes images, curve logging, geologic analyses and engineering parameters into account, instead of only statistic averages of log curves. The mixed data provides more aspects of features. In particular, structural images and curves contain more details and are fit for exploiting cumulative effect within reservoir formations. The HDNN model was applied to an oil development block to predict production, with an MLP model and a CNN model as comparisons. Results highlighted the superiority of HDNN model over the MLP or CNN model, with better optimization convergence and generalization performance. In addition, the performance of the HDNN and CNN model demonstrated that structural logs are especially suitable for productivity evaluation. Further application to three new wells for production prediction validated the feasibility of HDNN model in practice. It can be used to predict production potential of target formations or target wells, and further guide engineering operation of oil exploitation. In conclusion, the proposed HDNNs is conducive to perform incisive data mining, especially for tasks providing mixed inputs.

## Acknowledgement

The authors acknowledge the support and permission of PST Service Corporation to publish this paper. We also thank J. Qiu for valuable discussions on production characterization of the adopted oil block.